\documentclass[11pt]{article}

\usepackage[final]{acl}

\usepackage{amsmath}
\usepackage{amsfonts}
\usepackage{hyperref}
\usepackage{url}
\usepackage{booktabs}
\usepackage{multirow}
\usepackage{pgfplots}
\usepackage{graphicx}
\usepackage{xspace}
\usepackage{array}
\pgfplotsset{compat=1.18}
\usepackage{arydshln}
\usepackage{todonotes}
\usepackage{subfigure}

\newcommand{\model}{\textsc{Ewe}\xspace}
\newcommand{\lf}{LongFact\xspace}
\newcommand{\fava}{\textsc{Fava}\xspace}

\newcommand{\alpaca}{AlpacaFact\xspace}
\newcommand{\alpacaE}{AlpacaEval\xspace}
\newcommand{\bio}{Biography\xspace}
\newcommand{\vs}{VeriScore\xspace}
\newcommand{\safe}{SAFE\xspace}
\newcommand{\contriever}{Contriever\xspace}
\newcommand{\nest}{\textsc{Nest}\xspace}
\newcommand{\fs}{FActScore\xspace}
\newcommand{\cove}{\textsc{CoVe}\xspace}
\newcommand{\dragin}{\textsc{DRAGIN}\xspace}

\newcommand{\cut}[1]{}

\title{Improving Factuality with Explicit Working Memory}

\author{Mingda Chen\quad Yang Li\quad Karthik Padthe\\\textbf{Rulin Shao}\quad\textbf{Alicia Sun}\quad\textbf{Luke Zettlemoyer}\quad\textbf{Gargi Ghosh}\quad\textbf{Wen-tau~Yih}\\
Meta FAIR \\
\texttt{mingdachen@meta.com}}

\begin{document}

\maketitle

\begin{abstract}
Large language models can generate factually inaccurate content, a problem known as hallucination.
Recent works have built upon retrieved-augmented generation to improve factuality through iterative prompting but these methods are limited by the traditional RAG design.
To address these challenges, we introduce \model (Explicit Working Memory), a novel approach that enhances factuality in long-form text generation by integrating a working memory that receives real-time feedback from external resources. The memory is refreshed based on online fact-checking and retrieval feedback, allowing \model to rectify false claims during the generation process and ensure more accurate and reliable outputs. 
Our experiments demonstrate that \model outperforms strong baselines on four fact-seeking long-form generation datasets, increasing the factuality metric, \vs, by 2 to 6 points absolute without sacrificing the helpfulness of the responses.
Further analysis reveals that the design of rules for memory updates, configurations of memory units, and the quality of the retrieval datastore are crucial factors for influencing model performance.
\end{abstract}
\section{Introduction}
\label{sec:intro}

\begin{figure*}[t]
    \centering
    \includegraphics[width=0.8\textwidth]{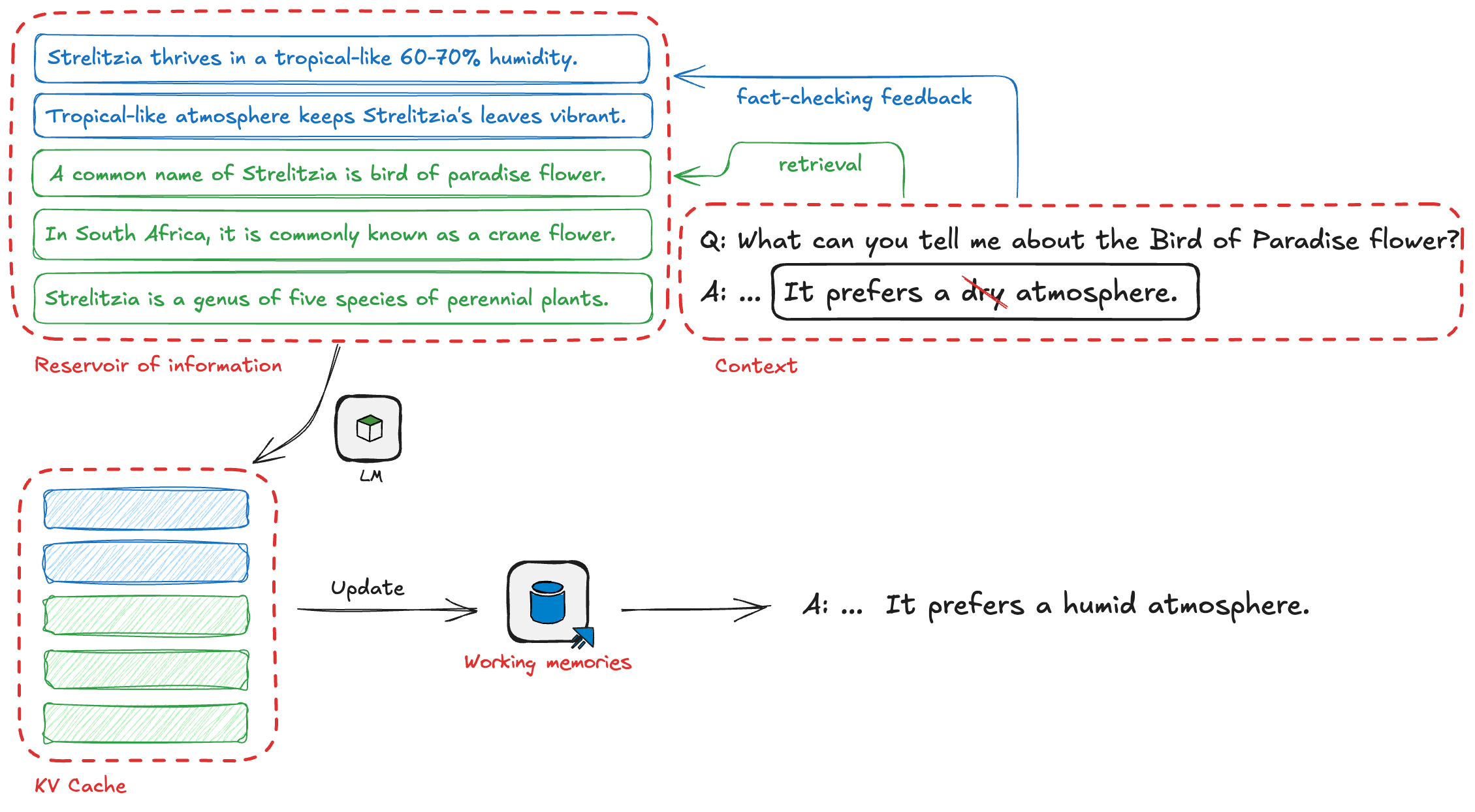}
    \caption{Example pipeline illustrating how \model pauses, receives feedback from retrievers and fact-checkers, and then re-generate to correct factual errors in its outputs. \model handles knowledge from fact-checkers and retrievers separately as they tend to provide information with distinct properties. Retrieval offers more general background information, while fact-checkers focus more on specific details, targeting particular aspects of the output text.}
    \label{fig:intro}
\end{figure*}

In the realm of long-form text generation, a notable vulnerability of large language models (LLMs) is their propensity for hallucination, wherein the generated text contains factually inaccurate information.
By prepending the input prompt with relevant documents from trustworthy sources, retrieved-augmented generation (RAG)~\citep{lewis-etal-2021-rag,shi-etal-2024-replug} has been shown to be a simple yet effective approach that substantially mitigates the hallucination issue.
To further enhance the factual accuracy of model output, various iterative prompting methods have been proposed that build upon RAG. For instance, FLARE~\citep{jiang-etal-2023-active} generates responses sentence by sentence, and if a newly generated sentence contains low-probability tokens, it retrieves a new set of documents and re-runs RAG to regenerate the sentence. Alternatively, Self-RAG~\citep{asai2024selfrag} employs a self-critic component to verify the correctness of each partial generation and repeatedly queries a retrieval system to update the background knowledge, thereby producing more accurate and faithful responses.
While these systems demonstrate significant empirical improvement, they are restricted in the traditional RAG design. Context-relevant knowledge through retrieval is the only online feedback to the model, incorporated as part of the input string.

In this work, we propose \model (\textbf{E}xplicit \textbf{W}orking m\textbf{E}mory), an iterative framework that aims to provide more factual responses for knowledge-intensive long-form generation, with the help of an auxiliary fact-checking module. \model augments an existing language model with an explicit working memory, which keeps track of the knowledge that is most relevant and useful at the current generation timestep. The memory is initially filled with the latent representation of some retrieved passages relevant to the input prompt. During the generation process, \model actively monitors the newly generated partial response and pauses occasionally to refresh the memory with knowledge from retrieval and to 
check the output statement. If the statement is factually incorrect, it then refreshes the memory with the fact-checking feedback. With the updated memory, \model first removes the incorrect statement and backtracks to the previous timestep, and then continues the generation process from there.

We assume that the main text generation model used here is a Transformer-based large language model, such as Llama~\citep{dubey2024llama3herdmodels}. Similar to the standard RAG setting, given an input prompt, we first retrieve $k$ relevant text chunks of the same number of tokens, as the background knowledge. Unlike RAG, which directly prepends the input prompt with these $k$ chunks, we apply the language model to them separately and store the KV cache of each chunk in a memory of $k$ units. When predicting the next token, the language model effectively attends the current token to all $k$ chunks in parallel, using their KV caches stored in the memory, and have the average as the attention value. 
When \model pauses the generation and checks the newly generated, partial response, it has the opportunity to update the memory in multiple ways to guide the language model. For instance, if some claims in the new sentence are not supported, this feedback along with additional supporting documents can be used as a new unit appended to the existing memory. In addition, if the knowledge from an initial retrieved passage is no longer relevant, its corresponding memory unit can be removed or updated with embeddings of a new passage retrieved using the generated partial response as query. 

\model can be seen as a more general framework that subsumes many existing approaches. For example, if there is no stopping in generation and if the memory contains only one unit (i.e., $k$=1), then \model degenerates to the simple vanilla RAG. If \model pauses at the end of generation of every sentence and checks whether the new sentence contains any token with low probability as a proxy of factuality measure, then this particular instantiation, with one memory unit, is effectively FLARE. Notice that in a typical, more general configuration of \model, the memory module consists of multiple units. When the memory is refreshed, not all the units need to be updated. If some knowledge is still required, their original raw data (e.g., passages) will not be reprocessed to create the embeddings, saving some redundant computational cost at inference time.
While conceptually simple, the working memory design in \model provides a more flexible and yet efficient way to incorporate various types of external online information, as different forms of feedback are encoded in parallel and stored in memory units (e.g., see Figure~\ref{fig:intro}). It is worth noting that \model is well-suited for real-time applications involving streaming user data, iterative retrieval, and real-time feedback. For example, when users input data through speech, where edits to previously entered information are less likely to occur, \model allows us to start gathering relevant information and drafting responses concurrently.
We notice that the design of leveraging working memory is also very related to some recently proposed methods for long-content models (e.g., Memory$^3$~\citep{yang2024text}). If our memory is used only for encoding the knowledge from the passages in our corpus, then this can be viewed as the whole corpus is used as the context, along with the prompt, as the input to the model. The key differences are that \model does not pre-encode every passage (although the KV caches of some frequently retrieved passages can certainly be precomputed in advance) and its memory can be dynamically updated as the generation progresses.

We demonstrate empirically that \model generates more factual responses without sacrificing the relevance to the input questions, using four fact-seeking long-form generation datasets. In general, with the feedback from online fact-checking and targeted retrieval, \model increases \vs~\citep{song-etal-2024-veriscore}, the factuality metric we use, by 2 to 6 points absolute and is equally helpful in terms of instruction following, compared to the base model Llama-3.1$_\text{70B}$. We also perform human evaluations to ensure that our enhancements in \vs effectively translate to factuality.

\section{Related work}
\label{sec:related}

Aiming to reduce hallucination and make the LLMs generate more factual responses, our proposed framework, \model, detects knowledge gaps and acquires relevant information as needed, incorporating feedback from auxiliary models when available.  Unlike chain-of-verification approaches~\cite[CoVe]{dhuliawala-etal-2024-chain}, which rely solely on the LLM for reasoning, \model combines adaptive retrieval augmentation and explicit memories with a focus on factuality. This section discusses related work on these two aspects. 

\subsection{Iterative and Adaptive Retrieval Augmentation}

Retrieval-augmented generation (RAG) typically involves a single retrieval step, followed by the language model generating a complete response. However, iterative retrieval methods~\cite[\S5]{gao2024retrievalaugmentedgenerationlargelanguage} have been proposed to generate responses in multiple segments, with each segment generated using different additional information retrieved through iterative retrieval.
One such approach is ITER-RETGEN~\citep{shao-etal-2023-enhancing}, which uses the model output of the previous iteration to formulate the query and retrieve more relevant knowledge for the current generation.
Extending iterative retrieval, the process of adaptive retrieval~\cite[\S5]{gao2024retrievalaugmentedgenerationlargelanguage} examines partially generated responses in previous iterations to decide whether retrieving new information or regenerating a segment response is needed. 
For instance, FLARE~\citep{jiang-etal-2023-active} follows a simple sentence-by-sentence generation process to answer fact-seeking questions. 
In each step, it generates a temporary next sentence and examines its acceptability based on model confidence. 
If the sentence is deemed questionable, it retrieves new text chunks using a query based on the temporary sentence and re-generates the next sentence using standard RAG.
DRAGIN~\citep{su-etal-2024-dragin} improves upon FLARE by introducing a new model confidence measure that combines attention scores and entropy values. This allows the model to pause the generation immediately after the confidence score of a token falls below a threshold. Additionally, DRAGIN uses preceding tokens with high attention scores on the stopping token to form a keyword-based query, which helps the model make a more confident next-token prediction.

Our work shares similarities with Self-RAG~\citep{asai2024selfrag}, particularly in the use of an auxiliary model to provide feedback. Unlike confidence measures based on token probability or attention score, Self-RAG fine-tunes the model to introspectively decide when to pause generation by outputting a special \texttt{retrieve} token. This triggers the retrieval of multiple passages, which are then used separately to generate candidate segments via standard RAG.
Each segment is evaluated by a ``critique'' model for relevance, usefulness to the original prompt, and support from the retrieved passage. The critique model's output determines whether a candidate segment is included in the final output. Similarly, previous research has investigated various methods for incorporating external feedback during the generation process~\citep{kang2023ever,gao-etal-2023-rarr,gou2024critic,li2024rac}.

Our approach, \model, differs from existing iterative and adaptive retrieval augmentation methods in two key aspects. Firstly, traditional retrieval augmentation is replaced with memory augmentation, where the representation is the KV cache (similar to TurboRAG~\citep{lu2024turborag}) instead of the raw text, and different memory chunks that encode different passages are processed in parallel. This design allows for greater flexibility in incorporating diverse information types and improves efficiency when only part of the memory is updated, as the remaining portion can be reused. Secondly, feedback from the auxiliary model is passed to the language model through memory, enabling the core language model to naturally incorporate multiple streams of information and produce better responses. This design difference sets our approach apart from existing methods and allows for more effective integration of factuality feedback from the auxiliary model.

\subsection{Memories in Long-context LLMs}

Incorporating a large-scale corpus as additional knowledge can be achieved by prepending the given prompt with all documents in the corpus as an extremely long context input~\citep{lee2024longcontextlanguagemodelssubsume} to language models. It is thus natural to see that long-context LLMs share some technical components that apply to retrieval augmentation.
The memory module in \model is analogous to the explicit memory design in Memory$^3$~\citep{yang2024text}.
Instead of encoding the knowledge in the training corpus completely in model parameters, or incorporating the knowledge primarily through retrieval augmentation, Memory$^3$ encodes 128-token chunks of the training corpus using their KV caches as memories. 
During inference, the model generates segments of 64 tokens. At the generation of each segment, it first uses the previous segment as query to retrieve 5 most relevant memories, and attends to them when generating the next segment.
Retrieving memories of KV caches has been proposed in earlier work. For instance, Memorizing Transformers~\citep{wu2022memorizing} effectively extends the context of the language model by $k$ nearest neighbor lookup of the past key-value pairs (i.e., long-range memory) and attends them in the last layer of the models. 
LongMem~\citep{Wang-augmenting-2023} proposed a decoupled network architecture, using the backbone language model as memory encoder and a trained residual side-network as memory retriever and reader. The top-$k$ attention key-value pairs stored in the memory are retrieved and incorporated at inference.

While we also use explicit memories to store KV caches in \model, our goal is to pass new information at each step in the iterative decoding process, such as new information relevant to the current context via online retrieval and feedback from auxiliary models. We allow different operations on existing memories, including update, append, or delete, providing more flexibility for various downstream tasks.

\section{Method}
\label{sec:method}

The overall generation process of \model is similar to the decoding process of typical Transformer-based models, with two differences: (\S\ref{sec:feedback}) \model pauses generation periodically. When a new complete sentence has been generated, \model uses the current context to retrieve a new set of passages as knowledge feedback. In addition, it runs a fact-checking model to judge whether the sentence contains any factually incorrect statements. 
If the sentence does contain factual errors, the correct facts will be used as the fact-checking feedback. Both types of feedback will be added to memories, and the sentence will be regenerated if the original one has factual errors. 
(\S\ref{sec:memories}) The generation is \emph{memory-augmented}. In addition to the typical context like the input sentence and tokens generated in previous timesteps, embeddings of various forms of feedback stored in the memories will influence the generated tokens through self-attention.

\subsection{Real-time Feedback}
\label{sec:feedback}

Following the design of recently proposed evaluation metrics on factuality~\citep{min-etal-2023-factscore,wei2024longform,song-etal-2024-veriscore}, we determine whether a sentence is factually correct by checking if all of the claims extracted from this sentence are supported.
While in general, \model can use any textual knowledge as feedback, we focus on providing two types of feedback when the newly generated sentence contains factual errors: \emph{fact-checking outcomes} and \emph{relevant knowledge}.

\paragraph{Fact-checking outcomes} This feedback consists of the correct information that refutes the inaccurate claims, such as \textit{``Strelitzia thrives in a tropical-like 60\%-70\% humidity.''} that proves \textit{``Bird of Paradise prefers a dry atmosphere.''} wrong in the example in Figure~\ref{fig:intro}. 
In this work, we adapt the claim extraction model and verification model in \vs~\citep{song-etal-2024-veriscore} as the fact-checking model, where the factual knowledge is derived from the Google snippets when using the extracted claim as query. 

\paragraph{Relevant knowledge} Using the original input question and the sentence being fact-checked as query, we use \contriever~\citep{izacard2022unsupervised} to retrieve passages from C4~\citep{JMLR:v21:20-074} and Wikipedia, following the setting in MassiveDS~\citep{shao2024scaling}. Passages with retrieval scores exceeding a certain threshold will be viewed as knowledge relevant to the current context and used to update the working memories.

We pause at every $T_r$ timesteps to gather feedback from retrievers, and $T_v$ timesteps from fact-checkers. However, if no new sentence is generated, the feedback collection process will be skipped.\footnote{\model is not significantly slower than existing iterative retrieval methods like \dragin and FLARE. These methods also use an iterative generation process involving re-generation and retrieval. However, without working memories, existing iterative retrieval methods are slower than EWE when it comes to incorporating new retrieved information into the context.}

\subsection{Refreshing Working Memories}
\label{sec:memories}

\begin{figure*}
    \centering
\includegraphics[width=0.6\textwidth]{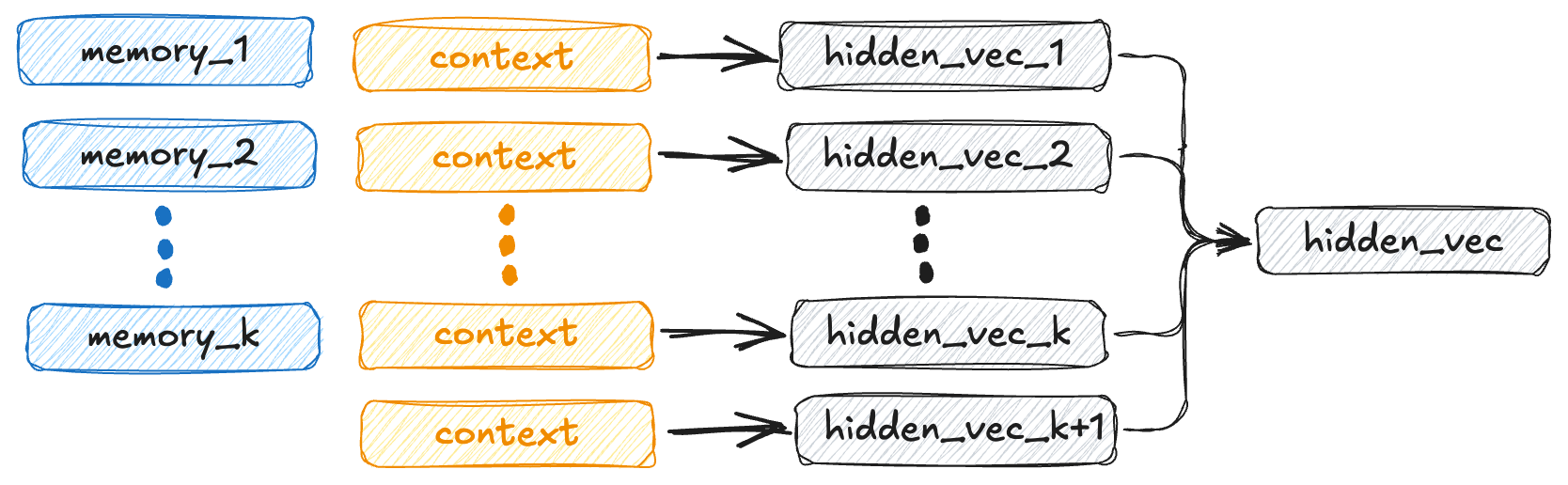}
    \caption{Diagram illustrating self-attention computations performed at each layer in \model. We concatenate each memory with the context (except for the last hidden vector where we only use context), apply standard self-attention and then aggregate the resulting hidden vectors to produce the final hidden vectors.}
    \label{fig:memory_arch}
\end{figure*}

The working memory in \model consists of $k$ memory units, where each unit is designed to store the representations of each feedback message of $M$ tokens. 
When updating working memories, we follow the \textit{first in, first out} (FIFO) rule.\footnote{In our preliminary experiments, we explored other possibilities of updating the memories, e.g., replacing the least relevant memories based on retrieval scores or model confidence scores (similar to what you mentioned). However, we found that these methods did not perform better than the FIFO approach. Due to the simplicity and effectiveness of FIFO, we decided to leave further exploration of these alternatives for future work.} 
Given refreshed text chunks of the same length from fact-checkers and retrievers, our model encodes them into the KV cache in parallel using the same positional IDs. Working memories in \model are stored as part of the language models' context preceding the model's own output text and prompts, allowing for flexible updates without reprocessing generated content. 
As shown in Figure~\ref{fig:memory_arch}, a separate embedding store is used for preserving these memories, which are then processed at each layer by concatenating them with the context. 
We then apply regular self-attention and aggregate the resulting hidden vectors using normalization terms from self-attention for each memory unit. Empirically, we find that adding hidden vectors produced by context only improves the fluency of long outputs, so we keep it in our model architectures. 
More formally, 
\begin{equation}
    \Vec{h}_n=\sum_{i=1}^{k+1}\frac{\alpha_i\Vec{h}_{n_i}}{\sum_{j=1}^{k+1}\alpha_j},
\end{equation}
\noindent where $\Vec{h}_n$ is the output vectors for self-attention at $n$-th layer in LMs, $\Vec{h}_{n_i}$ is the hidden vectors produced by memories concatenated with the context and by only the context vectors, and $\alpha_i$ is the normalization term from self-attention that leads to $\Vec{h}_{n_i}$.

\section{Experiments}
\label{sec:exp}

We present the main experimental results of \model in this section, along with details of the datasets and evaluation metrics we used, and the baseline we compared with. In this set of experiments, we set the retrieval and verification timesteps, $T_r$ and $T_v$, to be 1 and 8, respectively.

\subsection{Evaluation Datasets}
We evaluated \model and baseline models using four fact-seeking long-form generation datasets: \lf~(\citealp{wei2024longform}; 250 prompts), \fava~(\citealp{mishra2024finegrained}; 141 prompts), \alpaca~(\citealp{dubois2023alpacafarm,lin2024flame}; 241 prompts) and \bio~(\citealp{min-etal-2023-factscore}; 181 prompts). We include more detailed descriptions of each dataset in the appendix.

\subsection{Evaluation Metrics} 

We assess the quality of model responses to fact-seeking questions based on two key axes: \textit{factuality} and \textit{helpfulness}.
For evaluating factuality, we considered multiple automatic metrics, such as \fs~\citep{min-etal-2023-factscore} and \safe~\citep{wei2024longform}, but ultimately chose \vs~\citep{song-etal-2024-veriscore} as our primary evaluation metric. 
Although these metrics share a similar design that decomposes sentences into ``atomic claims'' and checks their support against an external knowledge source, \vs focuses on extracting more sensible verifiable claims and uses Google search snippets instead of Wikipedia as the knowledge source.
As a result, \vs can be applied to responses on more diverse topics and is also more efficient, requiring fewer but more meaningful claims to be checked. 
We report the F$_1$ score from \vs, which is the harmonic mean of the precision and recall of the claims. 
Following \citet{song-etal-2024-veriscore}, we set the minimum number of facts required for a model's response to achieve perfect recall as the median number of extracted claims per dataset\footnote{The median numbers of extracted facts for \lf, \fava, \alpaca, \bio are 55, 49, 31, 43, respectively.}. We also used their fine-tuned models for claim extraction and verification, provided in their package\footnote{\url{https://github.com/Yixiao-Song/VeriScore}}. 

To make sure that a model with a high factuality score does not simply give irrelevant but correct factual statements, we also need to check whether the response is helpful to the user.
Following \cite{lin2024flame}, we use \alpacaE~\citep{dubois2024lengthcontrolled} to compare the target model and baseline model in terms of their instruction-following ability. 
For the responses to the same input prompt, a large language model is used as judge to determine which of the two is better\footnote{We used GPT-4o as the judge.}, and the win rate is thus used as a measure of helpfulness\footnote{We found that the length-controlled win rates in \alpacaE could conflate hallucinations and length effects, and thus report the version without length normalization.}.

\begin{table*}[t]
    \centering
    \begin{tabular}{lcccccccc} \toprule
        \textbf{Model} & \multicolumn{2}{c}{\textbf{\lf}} & \multicolumn{2}{c}{\textbf{\fava}} & \multicolumn{2}{c}{\textbf{\alpaca}} & \multicolumn{2}{c}{\textbf{\bio}} \\
        & F$_1$ & WR & F$_1$ & WR & F$_1$ & WR & F$_1$ & WR \\\midrule
        Llama-3.1$_{\text{70B}}$ & 64.3 & - & 52.0 & - & 63.8 & - & 37.1 & - \\
        +RA & 72.7 & 41.2 & 56.8 & 37.1 & 66.0 & 43.1 & 43.8 & 49.4 \\
        +\nest  & 63.2 & 9.1 & 50.3 & 24.1 & 58.1 & 30.2 & 41.5 & 22.1 \\
        +\dragin & 71.5 & 38.2 & 57.2 & 33.9 & 65.3 & 31.5 & 42.8 & 33.5 \\
        +\cove & 63.8 & 39.3 & 49.5 & 33.4 & 61.5 & 33.3 & 37.7 & 31.3 \\
        +\cove w/ Retrieval & 67.4 & 31.8 & 52.6 & 23.1 & 64.0 & 28.8 & 38.2 & 29.4 \\
        +\model & \bf 75.9 & \bf 50.1  & \bf 61.0 & \bf 50.1 & \bf 66.9 & \bf 49.9 & \bf 49.7 & \bf 50.2 \\ \midrule
        Llama-3.1$_{\text{8B}}$ & 63.1 & \bf 40.6 & 51.0 & \bf 36.5 &\bf  65.3 & 26.7 & 28.9 & \bf 24.2 \\
        +RA & 65.9 & 28.1 & 51.8 & 16.8 & 63.9 & 18.5 & 41.4 & 21.3 \\
        +\nest & 62.3 & 4.2 & 50.2 & 14.1 & 57.8 & 9.1 & \bf 41.8 & 21.8 \\
        +\dragin & 63.9 & 15.9 & 51.1 & 10.0 & 61.3 & 11.1 & 34.7 & 11.4 \\
        +\cove & 44.1 & 8.8 & 38.7 & 11.0 & 51.3 & 15.1 & 25.1 & 13.3 \\
        +\cove w/ Retrieval & 53.5 & 12.2 & 39.5 & 5.3 & 54.6 & 12.5 & 29.1 & 10.2 \\
        +\model & \bf 67.3 & \bf 40.5 & \bf 53.1 & 36.2 & \bf 65.5 & \bf 28.0 & \bf 42.2 & 21.5 \\ \bottomrule
    \hline
    \end{tabular}
    \caption{Evaluation on factuality and helpfulness of the model responses to prompts provided in four long-form question answering datasets. For each dataset, we report F$_1$ scores from \vs and win rates (WR) from \alpacaE. We use Llama-3.1$_{\text{70B}}$ as the baseline method in \alpacaE win rate experiments.}
    \label{tab:main_result}
\end{table*}

\subsection{Baselines}

We used instruction-tuned Llama-3.1 70B and 8B as the base models\footnote{We opt to use Llama-3.1 in two different sizes instead of exploring different architectures. This decision is based on the fact that \model is not tied to any specific model architecture. Consequently, it is more insightful to assess its generalization capabilities across various model sizes rather than across different architectures.} and compared \model with five baselines: base model only, retrieval augmentation (RA), Chain of verification (\cove)\footnote{We adapted an implementation from \url{https://github.com/ritun16/chain-of-verification}}, an iterative retrieval approach \dragin~\citep{su-etal-2024-dragin}\footnote{We used the authors' implementation \url{https://github.com/oneal2000/DRAGIN}}, and a recently proposed semi-parametric decoding method \nest~\citep{li2024nearest}.
For base model only, Llama-3.1$_{\text{70B}}$ or Llama-3.1$_{\text{8B}}$, we simply gave the language model the prompt in the dataset and the instruction of requesting detailed information, without other additional information. 
With retrieval augmentation, we retrieved 10 passages using the input prompts as queries and then prepended the  passages to the input\footnote{Using more than 10 passages does not provide significant benefits in our preliminary experiments, so we limit our retrieval to the top 10 passages.}.
\nest is a strong retrieval-based decoding algorithm. Following the original setup, we retrieved 100 passages to use as candidates.
For \cove, we employ the ``factor+revise'' method, which \citet{dhuliawala-etal-2024-chain} demonstrated to be the most effective. Additionally, we improve \cove by integrating retrieved passages from our retrieval datastore during the verification step. This augmentation helps us establish a stronger and more comparable baseline method, considering that most other baseline methods also utilize retrieval.
For all our experiments, the maximum generation step was set to 1024. We also use C4 and Wikipedia in MassiveDS as our retrieval datastore and Contriever as the retriever. 
Llama-3.1$_{\text{70B}}$ is used as the baseline method for all \alpacaE comparisons.

\subsection{Results}

Our main results are shown in Table~\ref{tab:main_result}. For the Llama-3.1$_{\text{70B}}$ base model, we find that in terms of factuality, retrieval augmentation generally improves the results consistently across different datasets. 
This is expected as for fact-seeking prompts, specifically conditioning generation on relevant factual knowledge has been demonstrated to be an effective way to mitigate hallucinations.
\nest performs better than the base model on the \bio dataset, but not on others, and it appears that the \vs F$_1$ is lower than the standard retrieval augmentation.
It might suggest that the configuration or hyperparameter settings of \nest need to be further optimized, as \nest was originally evaluated by \bio with Llama-2. \dragin performs similarly to RA, likely because their query formulation method is not optimized for long-form generation, resulting in less useful retrieved passages. Similarly, with \cove, we notice that it often produces shorter model responses, leading to significantly lower recall performance despite high precision, which results in a less favorable \vs F$_1$. While augmenting \cove with retrieval slightly alleviates this issue, it still lags behind.
Perhaps more interestingly, with online fact-checking feedback and refreshed knowledge from retrieval, \model achieves the highest \vs F$_1$ on all datasets. On the helpfulness of the responses, it appears that \alpacaE generally prefers the output from the base model, except for \model, where the win rates are roughly 50\%.

\begin{figure}
\centering
\includegraphics[width=0.3\textwidth]{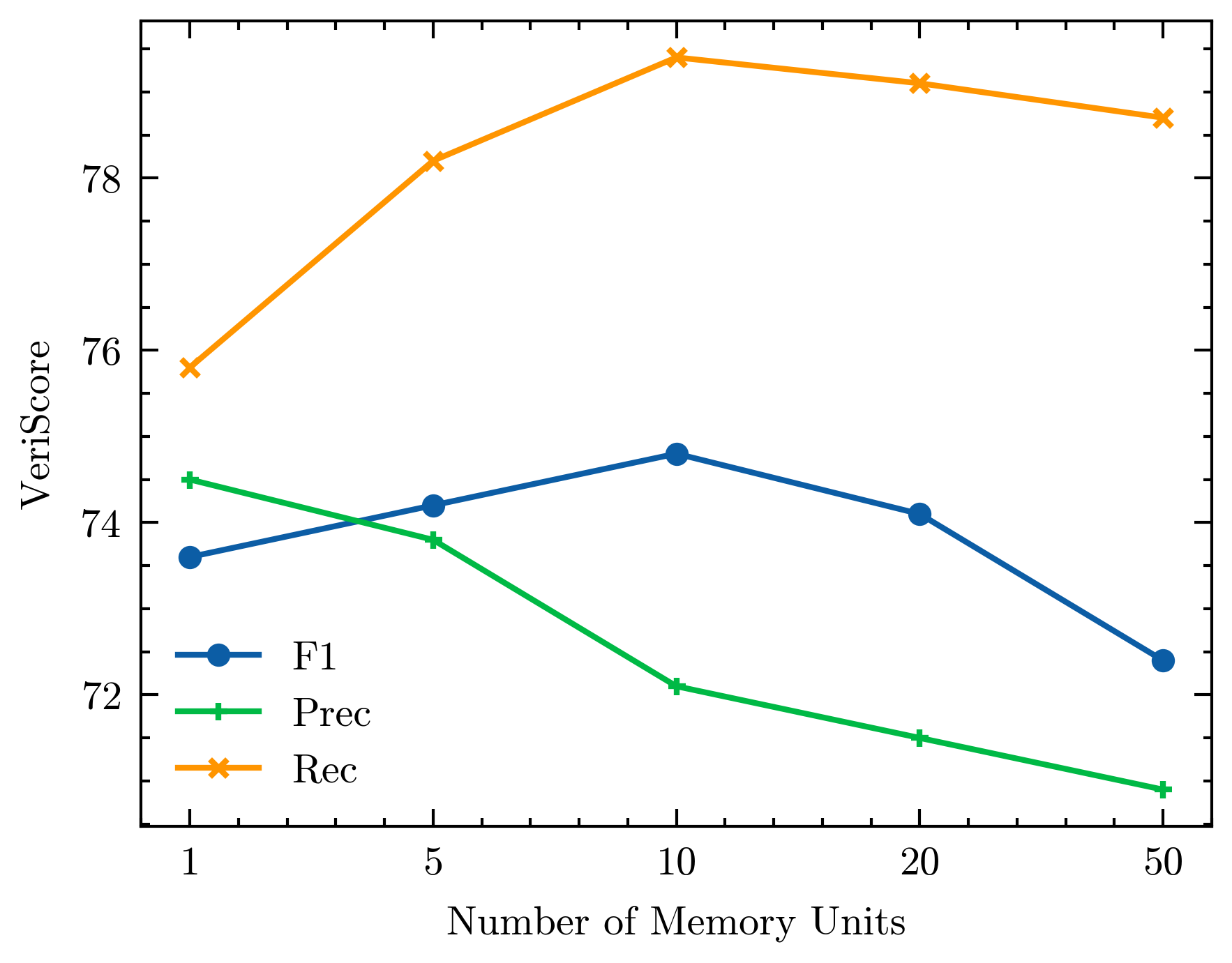}
    \caption{\vs over 50 prompts from \lf when varying number of memory units used for storing retrieved passages and fact-checking feedback.}
    \label{fig:memory_unit_number_ablation}
\end{figure}

When using Llama-3.1$_{\text{8B}}$ as the base model, we have observed a similar trend. Retrieval augmentation improves factuality in terms of \vs F$_1$ and \model still gives the best factuality results. However, compared to the models based on Llama-3.1$_{\text{70B}}$, we notice that the improvement is generally smaller.
We hypothesize that the smaller base language model is less capable in leveraging feedback, and may not always regenerate a sentence that is factually correct.
In terms of helpfulness, we can see that \model generally performs comparably to its base model Llama-3.1$_{\text{8B}}$, as they have similar win rates when judged against the output of the same Llama-3.1$_{\text{70B}}$ base model.  

\section{Analysis}
\label{sec:analysis}

We provide some insights based on different ablations in this section. 
Llama-3.1$_{\text{70B}}$ were used as the base model for all the experiments in this section. Due to space limits, we include analysis on model confidence, retrieval knowledge, and feedback forms for fact verifiers in the appendix.

\subsection{Memory Configurations}

In this analysis, we explore the influence of memory configurations on factuality, based on experiments on 50 randomly sampled prompts from \lf. We examine this impact through two dimensions: the number of memory units and the lengths of memory units.

In Figure~\ref{fig:memory_unit_number_ablation}, we investigate how varying the numbers of memory units used for storing fact-checking feedback and retrieved passages may impact factuality. We apply the same configuration to both, as similar trends are observed when adjusting the settings for each type of memory. Overall, we find that having a large number of memory units for either fact-checking feedback or retrieved passages \textit{negatively} affects precision and, consequently, factuality (though recall remains largely unaffected).
This is likely because a significant amount of stale information remains in working memory for an extended period without being updated, as we adhere to the FIFO rule for updating working memory. 
Consequently, this information becomes outdated as the generation process continues.

\begin{figure}
\centering
\includegraphics[width=0.3\textwidth]{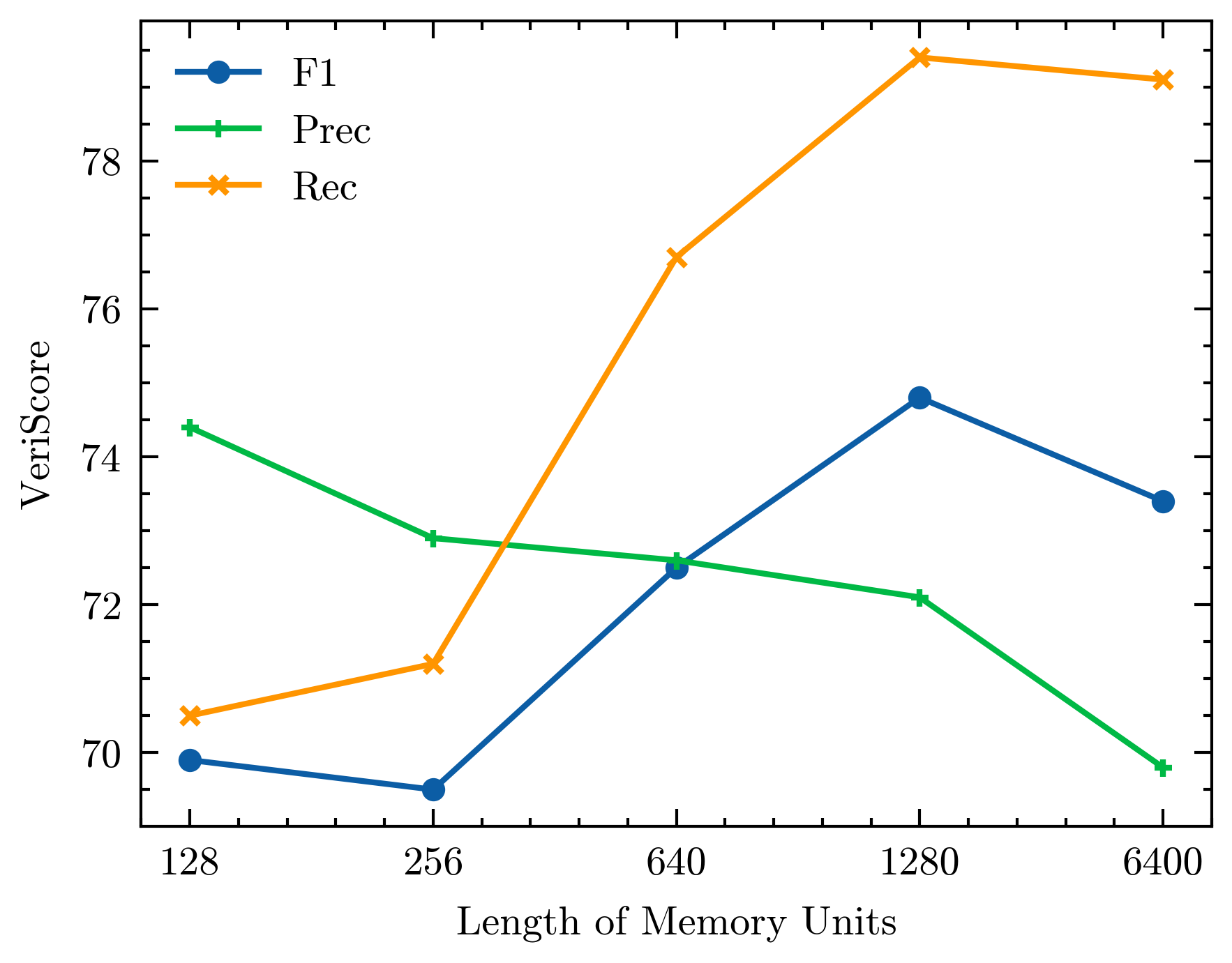}
    \caption{\vs over 50 prompts from \lf when varying lengths of memory units used for storing retrieved passages and fact-checking feedback.}
    \label{fig:memory_unit_length_ablation}
\end{figure}

In Figure~\ref{fig:memory_unit_length_ablation}, we examine the impact of varying lengths of memory units on factuality. Our results indicate that shorter memory units tend to increase precision but decrease recall, whereas longer memory units tend to decrease precision but increase recall. We hypothesize that this occurs because short memory units allow the attention mechanism to allocate weights more effectively to individual passages.
In contrast, longer memory units combine multiple passages into a single unit, which may compel the attention to focus on less relevant passages when they are grouped with more relevant ones. However, longer memory units also provide advantages in terms of improved reasoning across documents, which is beneficial for recall.

\begin{figure}
    \centering
\includegraphics[width=0.3\textwidth]{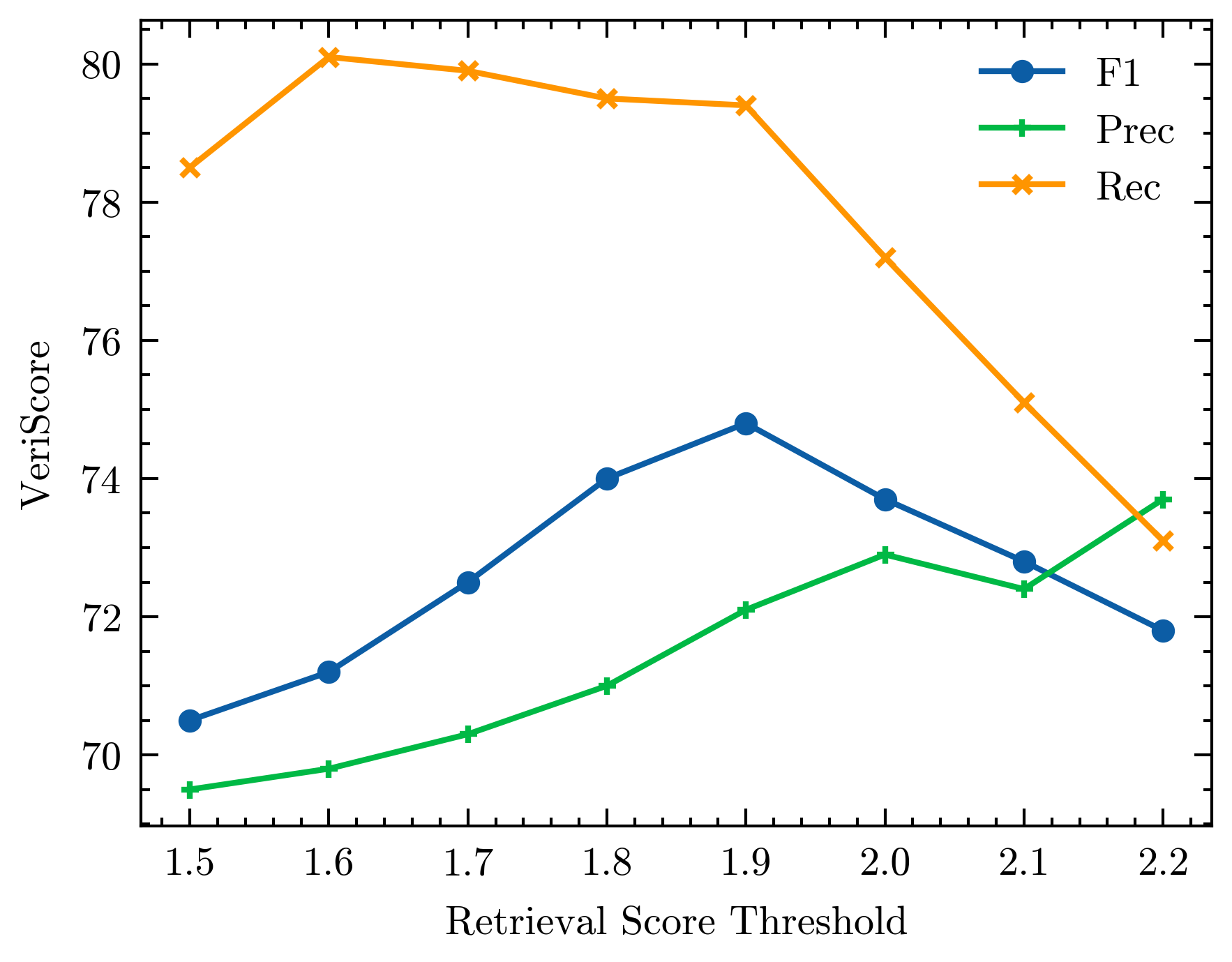}
    \caption{\vs over 50 prompts from \lf with different retrieval score thresholds. Higher thresholds indicate passages that are less relevant would be excluded in working memory.}
    \label{fig:retrieval_score_threshold_ablation}
\end{figure}

In Figure~\ref{fig:retrieval_score_threshold_ablation}, we examine the impact of varying retrieval score thresholds. Notably, initially raising the threshold (allowing more relevant information to be retained in working memory) enhances both precision and recall. Further increasing the threshold continues to improve precision while maintaining recall at a similar level. However, increasing the threshold beyond this point negatively affects F$_1$ and recall, despite still benefiting precision, likely due to the exclusion of too many retrievals. These findings indicate that our model might gain from a more advanced memory updating mechanism, which we intend to investigate in future work.

\subsection{Human Evaluation}
\begin{table}[t]
    \centering
    \begin{tabular}{l|c}\toprule
       RA  & 0.61 \\
       \model  &  0.65 \\\bottomrule
    \end{tabular}
    \caption{Cohen's Kappa inter-annotator agreement between \vs and human predictions regarding whether sentences are factual or not.}
    \label{tab:inter-annotator-agreement}
\end{table}

We conduct a human evaluation to determine if using \vs during inference impacts its alignment with human judgments. To achieve this, we randomly select 100 generation samples each from our retrieval-augmented baseline and \model, and recruit 6 human annotators. During sample selection, we categorize sentences into buckets based on the number of nonfactual claims predicted by \vs. We then randomly choose samples from each bucket to ensure an equal representation of sentences with varying numbers of nonfactual claims. Annotators are given a sentence from model responses along with the full response for context. They are instructed to classify the sentence as factual, nonfactual, or ``unable to decide'' if no claim is present. Additionally, they are encouraged to use online search engines if they are unsure about specific information. Detailed instructions are provided in the appendix. Our evaluation gathers 120 annotations. Since annotators might choose ``unable to decide'' for various reasons, such as the sentence being functional and not making any claims, we exclude these from score computation. We calculate Cohen’s Kappa to assess inter-annotator agreement between humans and \vs, excluding 25 ``unable to decide'' annotations. The results are presented in Table~\ref{tab:inter-annotator-agreement}. Since a Cohen's Kappa score above 0.61 is considered substantial and the scores do not decrease with \model, we conclude that using \vs during inference has indeed enhanced the factuality of \model.

\section{Conclusion}
\label{sec:conclusion}
We present \model, a novel system that incorporates a working memory mechanism during the generation process. 
\model pauses at given intervals and refreshes its working memory based on feedback from retrieval and fact-checking models, ensuring that the generated content remains accurate and relevant.
Our experiments demonstrate the effectiveness of \model by benchmarking it on 8B and 70B Llama-3.1 models, resulting in significant improvements in both factuality and helpfulness across four fact-seeking long-form generation datasets. 
Furthermore, our analysis reveals that updating the working memory with more relevant information at each timestep, allowing attention to focus on each passage, and utilizing high-quality retrieval datastores with extensive knowledge coverage are crucial factors for improving factuality of models.
\section{Limitations}
Although \model can be applied to various types of data, our research has been limited to English text datasets with fact-seeking prompts. Its performance on other English benchmarks and multilingual datasets remains unclear. Additionally, we have restricted \model to only accept textual real-time feedback, and exploring multimodal feedback could be intriguing. Moreover, our human evaluation experiments have been conducted on a relatively small scale, and it would be beneficial to validate our findings in a larger-scale setting.

\bibliography{paper,anthology_0,anthology_1}
\appendix

\section{Evaluation Datasets}
\paragraph{\lf} Designed to probe the factuality of a model of which response consists of at least several paragraphs, \lf was created by prompting GPT-4 to generate questions regarding a specific concept or object within a given topic.
In our experiments, we use the 250 prompts from the \lf-Objects dataset, selected by \citet{wei2024longform}.  
\paragraph{\fava} As a new fine-grained hallucination benchmark, \fava constructed 200 information-seeking queries that require factual knowledge to give accurate long-form answers from multiple sources, including Open Assistant~\citep{Kopf:OpenAssistant23}, No Robots~\citep{no_robots}, WebNLG~\citep{webnlg} and instructions written by the authors~\citep{mishra2024finegrained}. Following \citet{lin2024flame}, we selected 141 prompts from this collection for our experiments.
\paragraph{\alpaca} Originally collected from real-world interactions with various users, the 805 instructions in AlpacaFarm~\citep{dubois2023alpacafarm} was used for evaluating the instruction-following ability of different LLMs.
To focus our evaluation on factuality, we used a subset of 241 fact-seeking instructions selected by \citet{lin2024flame} in this work.
\paragraph{\bio} To demonstrate the effectiveness of the factuality metric \fs, \citet{min-etal-2023-factscore} selected 183 names of famous people found in Wikipedia, and applied the ``\textit{Tell me a bio of} [\texttt{Person Name}]'' template to create a collection of prompts called \bio. As this set of prompts have been used extensively in several recent papers, we include it in our study as well.

When using these prompts, we appended the instruction ``\textit{Provide as many specific details and examples as possible (such as names of people, numbers, events, locations, dates, times, etc.)}'' to encourage models to generate more detailed responses that cover multiple factoids, following \citet{wei2024longform}.

\section{Analysis on Model Confidence}

One important question remains is when to refresh the working memory. To study it, we conducted a comparative analysis of different criteria for refreshing working memory and regeneration. 
Since the working memory consists of the retrieval memory and fact-checking memory, which can have interacting effects, we first investigate when to trigger the retriever alone (without fact-checking memory) and then investigate when to trigger the fact-checker (when retrieval interval $T_r$ is set to 1).

\begin{figure}[t!]
\centering
\begin{subfigure}
        \centering
        \includegraphics[width=0.4\textwidth]{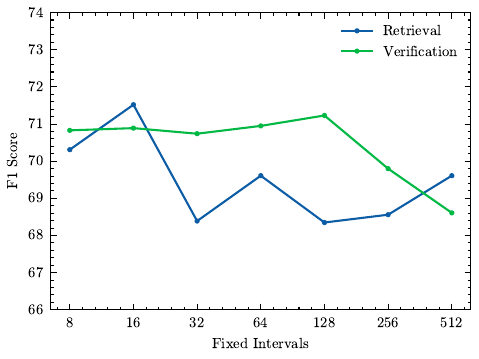}
        \caption{\vs F$_1$ when using fixed retrieval and verification intervals.}
        \label{fig:interval-fixed}
    \end{subfigure}%
    \begin{subfigure}
        \centering
            \includegraphics[width=0.4\textwidth]{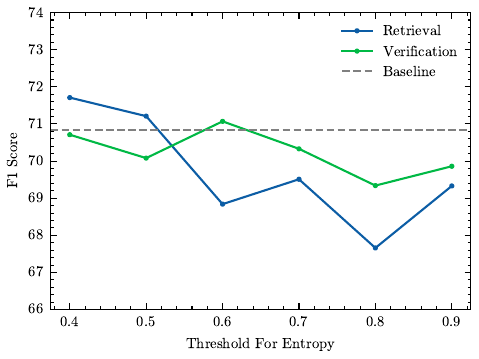}
        \caption{F$_1$ when using entropy thresholds for triggering retrieval and verification.}
                \label{fig:interval-entropy}
    \end{subfigure}
    \begin{subfigure}
        \centering
            \includegraphics[width=0.4\textwidth]{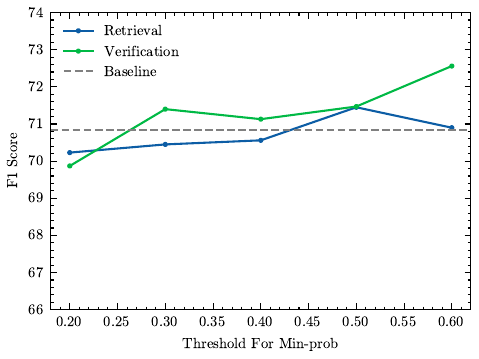}
        \caption{F$_1$ when using min-prob thresholds for triggering retrieval and verification.}
            \label{fig:interval-prob}
    \end{subfigure}
\caption{Comparison of different criteria for refreshing working memory over 50 prompts from \lf. The baseline uses retrieval interval $T_r = 1$ and verification interval $T_v = 8$.}
\label{fig:interval}
\end{figure}

\paragraph{Fixed intervals for refreshing working memory}
As shown in Figure~\ref{fig:interval-fixed}, when using a fixed retrieval interval, an smaller interval seems to perform better in general with a few spikes. 
This may be due to the fact that overly frequent retrieval can add irrelevant and conflicting information to the memory. On the other hand, when using a fixed verification interval, shorter interval always leads to better performance. 

\paragraph{Model confidence for refreshing working memory}
In practice, fixed retrieval and verification intervals may be unnecessary and lead to sub-optimal performance. We explore whether model-confidence can serve as a signal for refreshing working memory. Specifically, we compare two different metrics for model confidence: (1) \textbf{Entropy}: average entropy of generated tokens in a sentence, and (2) \textbf{Min-prob}: minimum probability of tokens in a sentence. A higher threshold for entropy results in less frequent memory update, and a higher threshold for min-prob results in more frequent memory update. 
Since external fact-checkers can be computationally expensive, we first examine if we can use model confidence as a signal for retrieval and regeneration, without using an auxiliary fact-checking model to provide feedback. 
As shown in Figure~\ref{fig:interval-entropy} and \ref{fig:interval-prob} (blue line), we observe empirically intermediate thresholds for retrieval perform well, leading to to better F$_1$ when compared to the settings in Figure~\ref{fig:interval-fixed}, where we use different fixed intervals for retrieval. 
With external fact-checkers, we investigate if we can use model confidence as a signal to trigger verification and regeneration to improve generation efficiency. 
In Figure~\ref{fig:interval-entropy} and \ref{fig:interval-prob} (green line), when chosen at an appropriate threshold, both entropy and min-prob can outperform the baseline (using fixed verification interval $T_v=8$ with the same memory configuration) despite with less frequent verification.

\section{Knowledge from Retrieval}

\begin{table*}[t!]
    \centering
    \begin{tabular}{lcccc} \toprule
       Datastore  & \lf & \bio & \alpaca & \fava \\\midrule
       Wiki      & 67.9 & 46.1 & \bf 55.5 & 52.5 \\
       C4        & 70.8 & 44.6 & 53.7 & \bf 53.3 \\
       C4 + Wiki & \bf 74.8 & \bf 48.4 & 53.3 & 52.3 \\ \bottomrule
    \end{tabular}
    \caption{\vs F$_1$ over 50 prompts from \lf, \alpaca, \fava and \bio with different retrieval datastores.}
    \label{tab:retrieval_corpus_ablation}
\end{table*}

We present the results of using different retrieval corpora in Table~\ref{tab:retrieval_corpus_ablation}, including Wikipedia, C4, or both of them together. Interestingly, different datasets exhibit distinct preferences. For \lf and \fava, C4 proves more effective than Wikipedia in helping the model to generate more factual responses. Conversely, \bio and \alpaca show a preference for Wikipedia.
Combining C4 with Wikipedia further improves factual accuracy for \lf and \bio, but the same trend is not observed for \alpaca and \fava. This is likely because the first two datasets consist of domains where C4 and Wikipedia provide complementary knowledge, whereas the latter two do not.

\begin{figure}
    \centering
    \includegraphics[width=0.4\textwidth]{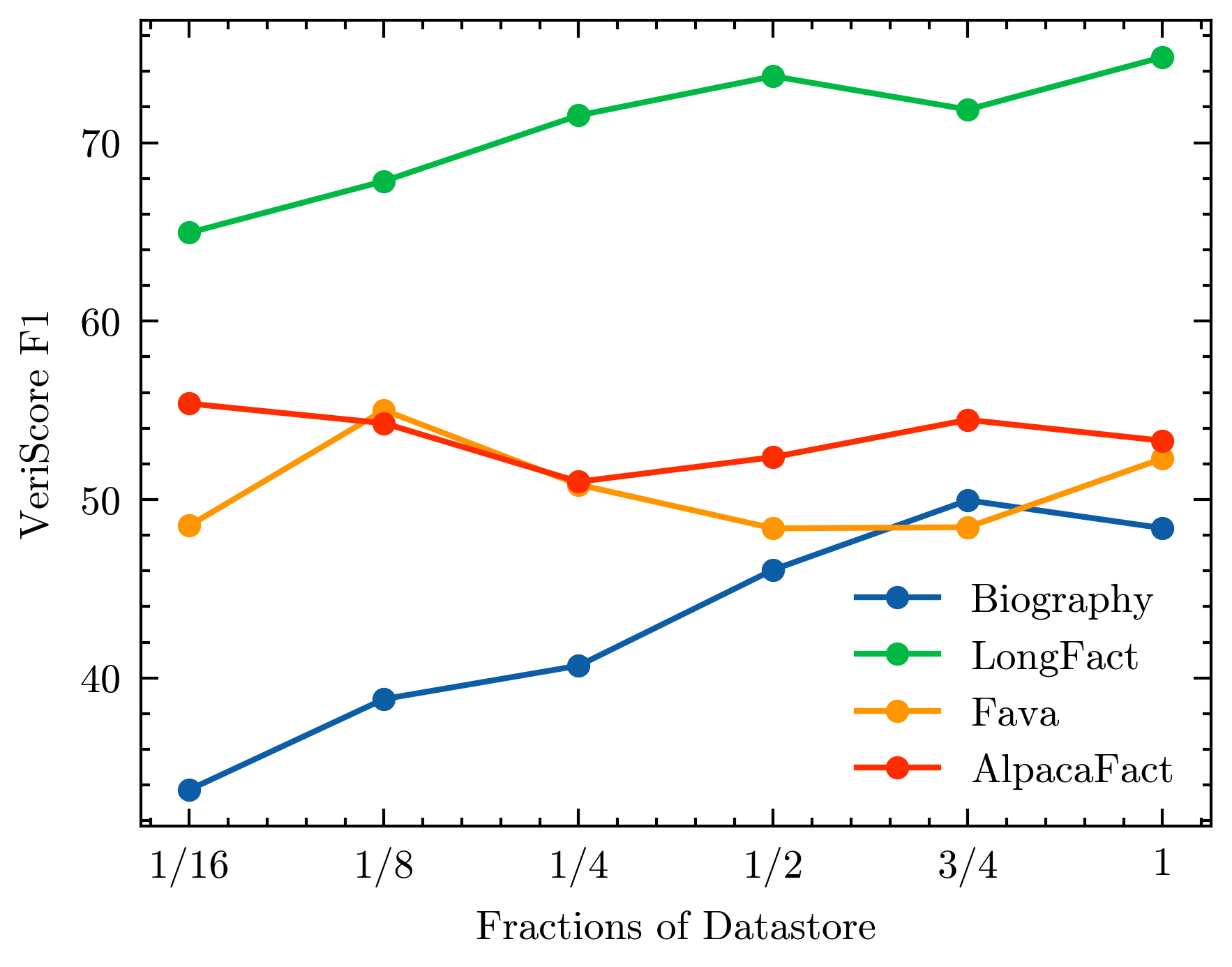}
    \caption{\vs F$_1$ when using varying fractions of datastore (C4+wiki).}
    \label{fig:retrieval_corpus_size_ablation}
\end{figure}

In Figure~\ref{fig:retrieval_corpus_size_ablation}, we explore the impact of using different sizes of retrieval datastores. Both \lf and \bio exhibit similar patterns, where increasing the datastore size generally improves results. In contrast, \fava and \alpaca differ, as \model tends to maintain similar level of \vs F$_1$ across various setups. We hypothesize that real-time feedback from fact verifiers helps offset the lack of information from retrieval datastores in these two datasets.

\section{Feedback Forms}

\begin{table*}[t]
    \centering
    \begin{tabular}{cccccc} \toprule
         \multirow{2}{*}{\begin{minipage}{1.5in}Passages determining a claim is incorrect\end{minipage}} & \multirow{2}{*}{\begin{minipage}{1.5in}Passages determining a claim is correct\end{minipage}} & \multirow{2}{*}{\begin{minipage}{1.5in}Instructions\\for nonfactual claims\end{minipage}}   & Precision & Recall & F$_1$  \\
        & & & & & \\\midrule
         \checkmark & \checkmark & \checkmark &  78.1 & 72.5 & 73.2 \\
         \checkmark & & & 77.1 & 73.5 & 72.2 \\
         & \checkmark & & 77.5 & 72.7 & 73.9 \\
         & & \checkmark & 71.1 & 76.1 & 72.5 \\
         \checkmark & \checkmark &   &\bf  72.1 & \bf 79.4 &\bf  74.8 \\
         & & & 70.8 & 76.0 &	73.0 \\\midrule
         \multicolumn{3}{r}{Llama-3.1$_{\text{70B}}$} & 65.8 &	67.1 &	65.5 \\
         \multicolumn{3}{r}{Llama-3.1$_{\text{70B}}$ + RA} & 70.7 & 75.6 & 72.5 \\ \bottomrule
    \end{tabular}
    \caption{Comparing different feedback forms for fact-checkers. We report \vs over 50 prompts from \lf.}
    \label{tab:feedback_form}
\end{table*}

In this analysis, we explore various feedback formats utilized by fact-checkers. The models in \vs offers 2 types of information: a list of both factual and nonfactual claims, along with relevant passages that support these factual and nonfactual judgments. To examine the impact of these feedback formats, we conduct experiments using different combinations of these information types in the working memory. For the supporting passages, we combine them using new line symbols. For the list of claims, we apply an instruction template as follows to encode nonfactual claims:
\begin{quote}
    Please refrain from including the following imprecise statements: (1) nonfactual claim\textsubscript{1} (2) nonfactual claim\textsubscript{2} ...
\end{quote}

Our results are shown in Table~\ref{tab:feedback_form}. Overall, fact-checking feedback is beneficial compared to the base model with and without retrieval augmentation. 
The specific types of feedback also play a crucial role. Incorporating all feedback forms does not enhance model performance, with supporting passages proving more effective than instructions. 
We notice that instructing models not to generate specific details often results in misunderstanding. Models might rephrase the instruction, include the nonfactual statement in their response, and then add a clarification indicating the previous statement is nonfactual, such as ``\textit{(Note: This is a nonfactual claim and may not be accurate.)}''. We leave a better design of feedback forms to future work. 
Interestingly, when we exclude all the textual feedback from fact-checkers and only pause and regenerate in the presence of nonfactual sentences, performance still slightly improves.

\section{Instructions for Human Annotators}
Below is the instruction we provide to each annotator.

\begin{quote}
    You will be given a sentence extracted from a model response, along with the full model response and the original prompt. Your task is to assess the factuality of the provided sentence. You may use Google search to help in your evaluation. A sentence should be considered as factual only if you can locate sources that corroborate all the claims made within it. If the sentence contains no claims, you may select "unable to decide."
\end{quote}

\end{document}